\pdfoutput=1

\documentclass[11pt]{article}

\usepackage{acl}

\usepackage{times}
\usepackage{latexsym}

\usepackage[T1]{fontenc}

\usepackage[utf8]{inputenc}

\usepackage{microtype}

\usepackage{inconsolata}

\usepackage{multirow, booktabs}
\usepackage{makecell, hhline}
\usepackage{xspace}
\usepackage{subfigure}
\usepackage{longtable}
\usepackage{amsmath,amssymb,mathtools,amsfonts,amsthm}

\DeclareMathOperator{\Corpus}{\mathcal{C}}
\DeclareMathOperator{\Matrix}{\mathcal{M}}
\DeclareMathOperator{\Prob}{\mathcal{P}}

\DeclareMathOperator{\Candidate}{\mathcal{S}}

\title{Large Language Models Know What Makes Exemplary Contexts}

\author{Quanyu Long\textsuperscript{\rm 1}~~ Jianda Chen\textsuperscript{\rm 1}~~ Wenya Wang\textsuperscript{\rm 1}~~ Sinno Jialin Pan\textsuperscript{\rm 1,2} \\
\textsuperscript{\rm 1}Nanyang Technological University, Singapore\\
\textsuperscript{\rm 2}The Chinese University of Hong Kong~~ \\
\texttt{\{quanyu001, jianda001, wangwy\}@ntu.edu.sg~~ sinnopan@cuhk.edu.hk}
}

\begin{document}
\maketitle

\begin{abstract}
In-context learning (ICL) has proven to be a significant capability with the advancement of Large Language models (LLMs).
By instructing LLMs using few-shot demonstrative examples, ICL enables them to perform a wide range of tasks without needing to update millions of parameters.
This paper presents a unified framework for LLMs that allows them to self-select influential in-context examples to compose their contexts; self-rank candidates with different demonstration compositions; self-optimize the demonstration selection and ordering through reinforcement learning.
Specifically, our method designs a parameter-efficient retrieval head that generates the optimized demonstration after training with rewards from LLM's own preference.
Experimental results validate the proposed method's effectiveness in enhancing ICL performance. Additionally, our approach effectively identifies and selects the most representative examples for the current task, and includes more diversity in retrieval\footnote{Codes of this work are available at \url{https://github.com/ruyue0001/RL-ICL}}.
\end{abstract}

\section{Introduction}
\label{sec:intro}
Recent advancements in Large Language Models (LLMs) through in-context learning (ICL) have demonstrated substantial capability across various tasks \citep{brown2020language}. 
Previous studies have shown that when the entire training dataset is available, employing a retrieval model to fetch semantically similar demonstrations significantly enhances the ICL performance~\citep{liu2022makes}. 
Consequently, numerous works have focused on enhancing dense retrievers to select more representative demonstrations based on the LLM feedback calculated through contrastive loss \citep{rubin2022learning,li2023unified,wang-etal-2024-learning}.

\begin{figure}[t]
    \centering
    \setlength{\abovecaptionskip}{0.2cm}
    \setlength{\belowcaptionskip}{-0.4cm}
    \includegraphics[width=0.99\columnwidth]{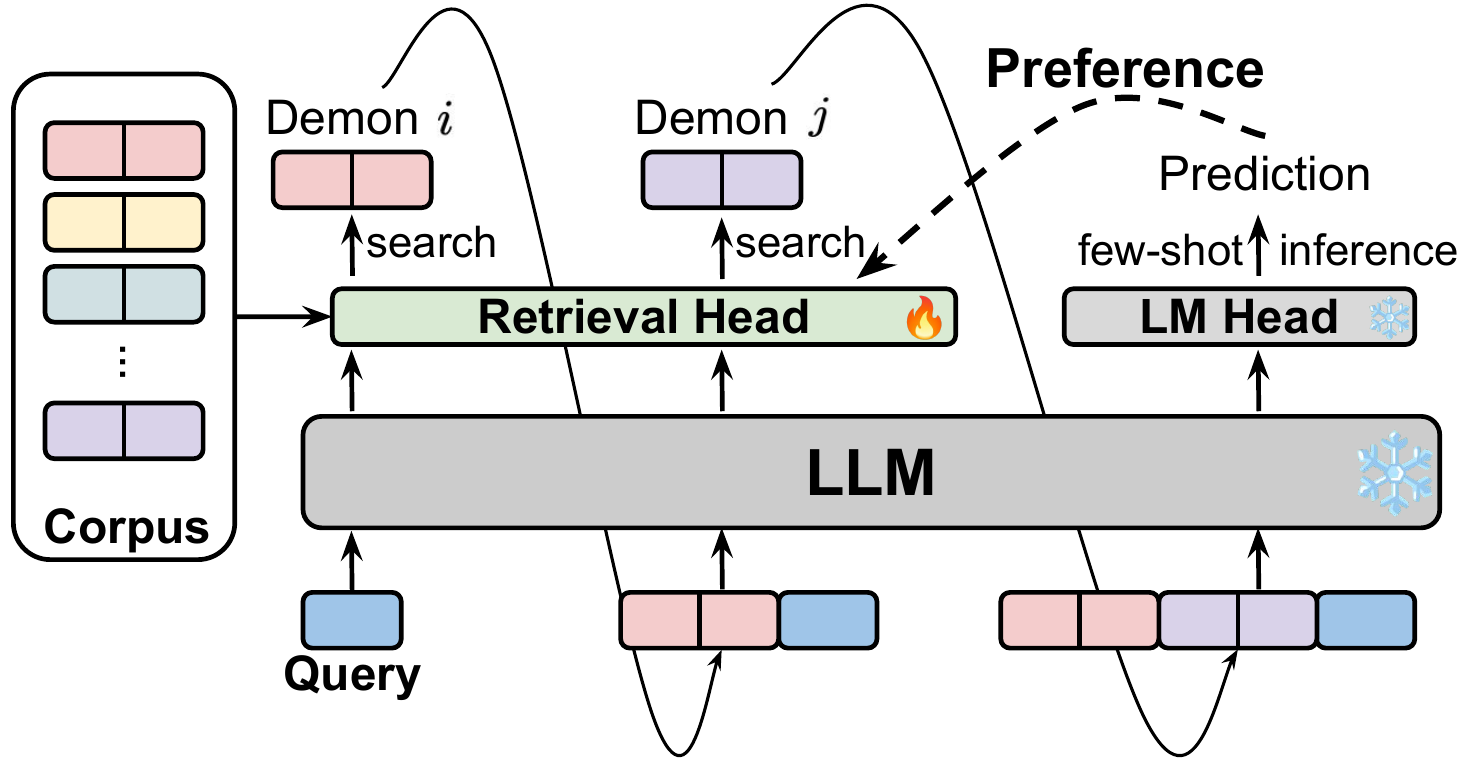}
    \caption{A self-select, self-rank, and self-optimize framework to retrieve influential in-context examples. 
    The LLM can select its own demonstrations sequentially by updating the contexts of query, and optimize their compositions based on the LLM's own preference.
    }
    \label{fig:moti}
\end{figure}

Despite significant advancements, there are still several challenges in this field.
Firstly, there exists a notable disparity between inference LLMs and encoder-based dense retrievers, which are generally Small Language Models (SLMs). Relying on SLMs to determine good in-context examples diminishes retrieval efficiency due to the discrepancy and limited capabilities of SLMs compared to LLMs themselves.
Secondly, previous works~\citep{rubin2022learning,li2023unified,wang-etal-2024-learning} rely on a single demonstration to gather LLM feedback during training but retrieve the top-$k$ results for testing. This discrepancy leads to suboptimal learning outcomes as the in-context examples have mutual influences. 
Lastly, the existing works fail to optimize the ordering of the given $k$ retrieved demonstrations,  which is crucial since ICL is sensitive to the order of demonstrations~\citep{lu-etal-2022-fantastically}.

\begin{figure*}[t]
\setlength{\abovecaptionskip}{0.1cm}
\setlength{\belowcaptionskip}{-0.4cm}
\centering
\includegraphics[width=0.99\textwidth]{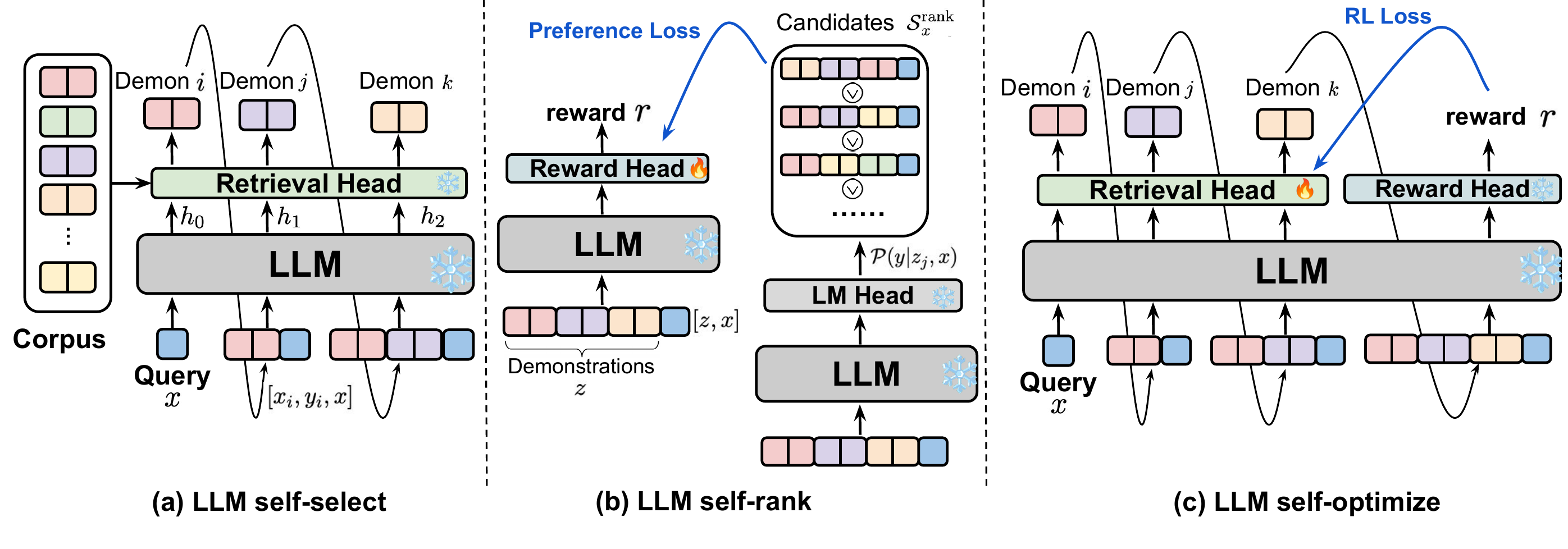}
\caption{Proposed Framework Overview. (a) sequential ICL retrieval formulation, (b) the first stage for training the reward head, and (c) the second stage for training the retrieval head.}
\label{fig:model}
\end{figure*}

To tackle the aforementioned challenges, we frame the ICL retrieval problem as a sequential retrieval process and employ reinforcement learning (RL) to model the interplay between exemplar selection and ICL inference as illustrated in Figure~\ref{fig:moti}. 
Technically, our method formulates a parameter-efficient retrieval head that generates an index of the most representative demonstration after training with rewards based on its preference from LLM responses.
Our approach offers several advantages:
\textbf{1) A single model does all}. We use a single, frozen LLM for both retrieval and inference, allowing the LLM to determine what makes its optimal context.
\textbf{2) Efficiency}. Unlike fine-tuning a dense retriever, our method only updates the retrieval head and reward head.
\textbf{3) Optimized mutual influence}. Our method selects demonstrations sequentially, optimizing their ordering and mutual influence.
\textbf{4) Enhanced diversity.} Recent studies underscore diversity also plays a critical role in selecting demonstrations~\citep{levy-etal-2023-diverse,long2024decomposing}. Our RL algorithm inherently includes more diversity through sampling.

\section{Method}
\label{sec:method}
\paragraph{Problem Definition}
Given an input $(x,y)$ and a corpus of in-context examples $\Corpus=\{(x_{i},y_{i})\}^{N}_{i=1}$, 
consisting of text-label pairs (demonstrations), 
our task is to select the $k$ most influential demonstrations from the corpus $\Corpus$ for the current query $x$. Additionally, we explore optimizing the order of these $k$ demonstrations. When prompting the LLM with $[z,x]=[x_{i_1},y_{i_1},...,x_{i_k},y_{i_k},x]$ ($[\cdot]$ denotes text concatenation, $z$ is $k$ stacked demonstrations) via few-shot in-context learning (ICL), we aim for the LLM to yield a high $\Prob(y|z,x)$.

\paragraph{Retrieval head modeling}
Instead of relying on a third-party retriever to retrieve top-$k$ most similar examples~\citep{liu2022makes,li2023unified,wang-etal-2024-learning}, we use the LLM itself to choose contexts. We introduce a matrix called \textbf{retrieval head} $\Matrix\in \mathcal{R}^{N\times D}$ to produce the policy distribution (selecting a demonstration from the corpus), where $D$ is the model dimension. The retrieval head is initialized using LLM-encoded sentence embeddings of $N$ demonstrations within the corpus $\Corpus$.

As shown in Figure~\ref{fig:model}(a), initially, the LLM encodes query $x$ to obtain the hidden state $h_{0}$. The logits at step 0 are calculated using $h_{0}\Matrix$, the $i$-th entry of the logits denotes the similarity score between the current state and the $i$-th demonstration in the corpus. The LLM selects the first demonstration based on the policy distribution $\pi_{\Matrix}(\cdot|h_{0})={\rm softmax}(h_{0}\Matrix)$. The query $x$ is then concatenated with the selected demonstration $(x_{i},y_{i})$ to produce the next-step input $[x_{i},y_{i},x]$. In step 1, similarly, the retrieval head selects the second demonstration to update the context. This auto-regressive process continues until $k$ demonstrations are gathered, resulting in the final ICL input $[z,x]$.

\paragraph{Reward model training}
Our designed reward model builds a 2-layer MLP (\textbf{reward head}) upon the LLM, 
which produces a scalar reward $r(\cdot)$ that estimates the quality of demonstrations (Figure~\ref{fig:model}(b)).
In contrast to prior studies that employ contrastive loss for grouping top and bottom-ranked candidates in reward modeling~\citep{li2023unified,wang-etal-2024-learning}, we instead use pair-wise preference to capture subtle differences between two high-ranked candidates.

Specifically, we first construct a candidate set $\Candidate_{x}=\{z_{j}\}^{m}_{j=1}$ by sampling $p_{t}$ demonstrations at each step $t$ in Figure~\ref{fig:moti}(a), with total number of candidates $m=\Pi_t^k p_t$.
Each candidate $z_{j}$ together with query $x$ is fed into the LLM and \textbf{LM head} that produces the probability score $\Prob(y|z_{j},x)$, where a higher probability of $y$ represents the demonstration combination $z$ is more favorable by LLM for current input $x$.
The candidate set $\Candidate_{x}$ is ranked in descending order of $\Prob(y|z_{j},x)$ indicating the LLM preference. 
We then train the reward model on the preference data using the Bradley-Terry model~\citep{bradley1952rank,NEURIPS2022_b1efde53}. 
The preference data is constructed according to the ranked candidate set $\Candidate_{x}^{\rm rank}$. We sample two ICL input $z^+$ and $z^-$ from $\Candidate_{x}^{\rm rank}$, and denote that $z^+$ has higher rank than $z^-$.
The preference loss of the reward model is defined as:
\begin{equation*}
    \mathbb{E}_{x} \left[\mathbb{E}_{(z^+,z^-) \sim \Candidate_{x}} \ \text{log} \sigma (r([z^+,x])-r([z^-,x])) \right],
\end{equation*}
where $\sigma(\cdot)$ is a sigmoid function.

\paragraph{Reinforced retrieval head from self-feedback}
To optimize the discrete demonstration selection process and maximize the LLM output posterior, 
we leverage the trained reward head
which provides stable rewards
to improve the retrieval head $\Matrix$ and retrieval policy $\pi_{\Matrix}$.
Similar to previous RLHF approaches~\citep{NEURIPS2022_b1efde53}, the objective for $\pi_{\Matrix}$ is regularized by a penalty on the KL-divergence between $\pi_{\Matrix}$ and $\pi_{\widehat{\Matrix}}$: 
\begin{equation}
\label{eq:rl_objective}
    \arg \max_{\Matrix} \mathbb{E} \ \left[r([z,x]) - \beta D_{KL}(\pi_{\Matrix} || \pi_{\widehat{\Matrix}}) \right],
\end{equation}
where $\pi_{\widehat{\Matrix}}$ is the reference policy, and $\widehat{\Matrix}$ is the initialized retrieval head. The KL-divergence regularizes the policy to remain close to the initial policy, as the reward model is trained on data sampled from the initial policy. The objective~\eqref{eq:rl_objective} is optimized via Proximal Policy Optimization (PPO)~\citep{SchulmanWDRK17} to update the retrieval head $\Matrix$.
As a result, the reinforced retrieval head produces an optimal policy of selecting the best demonstration composition, rendering enhanced ICL performance.


\section{Experiments}
\label{sec:exp}
{\renewcommand{\arraystretch}{1.05}
\begin{table*}[t]
\small
\centering
\setlength{\abovecaptionskip}{0.2cm}
\setlength{\belowcaptionskip}{-0.4cm}
\setlength\tabcolsep{4.5pt}
\begin{tabular}{lccccccccccc}
\Xhline{1.5pt}
 & SST-2 & MR & CR & ComE & ComV & RTE & SNLI & Reddit & Roc Ending & PHP & DART         \\
\Xhline{1pt} 

Random                          &88.87&82.10&79.49&42.30&62.40&60.80&40.78&19.12&23.08&28.77&37.73          \\
BM25                            &90.14&86.05&83.04&44.40&67.60&70.55&45.05&19.75&25.33&30.48&40.70         \\
SimCSE                          &91.97&\textbf{89.40}&85.82&51.20&67.10&71.48&51.52&19.27&26.25&32.20&40.98         \\
SBERT                           &90.45&86.35&84.24&53.30&68.30&70.46&52.40&19.31&24.97&31.44&39.09         \\
\citet{wang-etal-2024-learning} &92.31&87.45&87.33&56.80&70.20&72.34&\textbf{58.96}&19.85&26.18&33.89&41.47         \\ \Xhline{0.8pt}
Llama3                          &89.56&85.80&85.77&52.00&60.30&68.59&44.24&15.50&25.55&33.20&39.35         \\ 
RL-ICL (Ours)                 &\textbf{93.34}&87.90&\textbf{87.42}&\textbf{57.40}&\textbf{80.70}&\textbf{75.81}&54.14&\textbf{20.04}&\textbf{26.29}&\textbf{35.83}&\textbf{41.68}         \\
\bottomrule

\end{tabular}
\caption{Main results on classification and generation tasks.}
\label{tab:main}
\end{table*}
}

\paragraph{Datasets}
We conduct experiments on a wide range of NLP tasks, including Sentiment Analysis: \textbf{SST-2}~\citep{socher-etal-2013-recursive}, \textbf{MR}~\citep{pang2005seeing} and \textbf{CR}~\citep{kim2022beyond}; Commonsense Validation and Explanation: \textbf{ComE} and \textbf{ComV}~\citep{wang-etal-2019-make}; Natural Language Inference: \textbf{RTE}~\citep{dagan2005pascal,bar2006second,giampiccolo2007third,bentivogli2009fifth} and \textbf{SNLI}~\citep{bowman-etal-2015-large}; Text
Summarization: \textbf{Reddit}~\citep{kim-etal-2019-abstractive}; Story Generation: \textbf{Roc Ending}~\citep{mostafazadeh-etal-2016-corpus}; Code Summarization: \textbf{PHP}~\citep{lu2021codexglue}; Data to Text: \textbf{DART}~\citep{nan-etal-2021-dart}. More details are provided in Appendix \ref{sec:appendixA}.

\paragraph{Experiment Setup}

Throughout our experiments, we use Llama-3-8B as the base LLM, keeping it frozen and shared for demonstration retrieval and ICL inference. 
The reward head is configured as a 2-layer MLP with 8192 hidden units.
We set the number of demonstration $k=3$, i.e., the maximum demonstration decoding length is 3.
The candidate set size is $m=12$, with $p_t$ sampling number per step being $p_{t=[0,1,2]}=[3,2,2]$.
The KL-divergence coefficient $\beta$ is set to $1e-3$ during RL policy training via PPO.
The sentence embedding is computed via the mean pooling of hidden states.
We set the batch size 32, 100 epochs for reward model training and 10k learning steps for PPO training.
By fixing the parameters of LLM and enforcing the retrieval policy to stay close to the original policy through KL-divergence, we cache the hidden states for seen inputs $[z,x]$, this results in a 10× acceleration in RL training. Consequently, reward model training together with RL training complete within 4 hours on a single NVIDIA A100 GPU.

\paragraph{Baselines} 
All baseline comparisons utilize the same inference LLM, with variations in the retrieval model and method. 
Specifically, \textbf{Random} involves sampling $k$ demonstrations randomly from the training set. \textbf{BM25}~\citep{robertson2009probabilistic} is a sparse retriever to retrieve $k$ most similar demonstrations. For dense retrievers, we adopt \textbf{SBERT}~\citep{reimers-gurevych-2019-sentence} and \textbf{SimCSE}~\citep{gao2021simcse} to retrieve top-$k$ examples. \citet{long2024decomposing} employ an E5-based retriever~\citep{wang2022text} and iteratively train the retrievers based on the LLM feedback. \textbf{Llama3} is the initial sequential demonstration retriever with initial retrieval head $\widehat{\Matrix}$ (without reward model training or reinforcement learning). \textbf{RL-ICL} is the model trained with our proposed method. RL-ICL can retrieves its own favorable examples.

\paragraph{Main Results}
Table~\ref{tab:main} presents the main results across 11 tasks. From the table, we can observe that our method outperforms baselines on most tasks, and significantly surpasses the Llama3 baseline which can be regarded as step 0 of the RL training. These results indicate that our proposed sequential demonstration retriever is comparable with dense retrievers (Llama3 v.s. dense retrievers). Furthermore, the comparison between RL-ICL and Llama3 reveals that the proposed framework, which retrieves its in-context examples, can be effectively trained using self-preference and reinforcement learning algorithms.

\paragraph{Necessity of reward model}
To assess the impact of the reward model, we created a variant "w/o reward model," in which RL is trained using the LLMs' raw log probabilities instead of the reward model's estimates. As shown in Table~\ref{tab:RM}, this variant performs worse than RL-ICL and even underperforms the initial retriever Llama3 in two tasks, highlighting the necessity of a stable reward model for effective RL training.

{\renewcommand{\arraystretch}{1.05}
\begin{table}[t]
\small
\centering
\setlength{\abovecaptionskip}{0.2cm}
\setlength{\belowcaptionskip}{-0.4cm}
\setlength\tabcolsep{4.5pt}
\begin{tabular}{lccc}
\Xhline{1.5pt}
 & SST-2 & CR  & RTE     \\
\Xhline{1pt} 
Llama3                          &89.56 & 85.77 & 68.59  \\ 
w/o reward model                &89.98 & 84.32 & 66.78  \\
RL-ICL (Ours)                 &93.34 & 87.42 & 75.81  \\
\bottomrule  
\end{tabular}
\caption{Comparison to w/o reward model.}
\label{tab:RM}
\end{table}
}

\paragraph{Representativeness and diversity of retrieved demonstrations}
We introduce two metrics to evaluate the quality of retrieved demonstrations: \emph{representativeness} and \emph{diversity}. The \emph{representativeness} metric quantifies the ratio of selected demonstration to the entire corpus, where a lower percentage indicates a better selection of representative demonstrations for the target task. The \emph{diversity} dimension assesses the variety of classes among the $k$ retrieved demonstrations (higher is better). As discussed in \citet{long2024decomposing}, diverse labels enhance label space and format recognition capability, which is vital for ICL.
As shown in Table~\ref{tab:r_and_d}, the optimized retrieval head selects influential demonstrations from a smaller subset of the corpus while including more diversity, leading to improved ICL performance.

{\renewcommand{\arraystretch}{1.05}
\begin{table}[t]
\small
\centering
\setlength{\abovecaptionskip}{0.2cm}
\setlength{\belowcaptionskip}{-0.4cm}
\setlength\tabcolsep{4.5pt}
\begin{tabular}{lccc}
\Xhline{1.5pt}
 & & RL-ICL & SimCSE \\
\Xhline{1pt} 
\multirow{2}{*}{SST-2}  & representativeness          & 1.2\%  & 29.56\% \\
                        &diversity                      & 1.98  & 1.42 \\
\multirow{2}{*}{RTE}    & representativeness          & 8.0\%  & 38.79\% \\
                        &diversity                      & 1.84  & 1.73 \\
\bottomrule  
\end{tabular}
\caption{Representativeness and diversity of RL-ICL .}
\label{tab:r_and_d}
\end{table}
}

\begin{figure}[t]
    \centering
    \setlength{\abovecaptionskip}{0.2cm}
    \setlength{\belowcaptionskip}{-0.4cm}
    \includegraphics[width=0.9\linewidth]{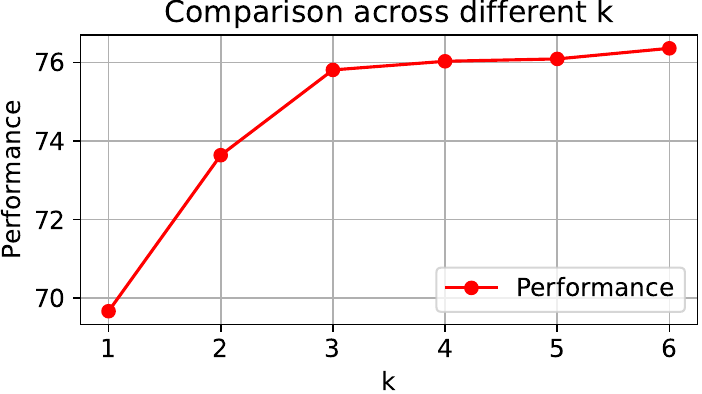}
    \caption{RTE performance using different $k$. }
    \label{fig:different_K}
\end{figure}

\paragraph{The number of demonstration $k$}
We evaluate the performance cross varying numbers of demonstrations $k$ from 1 to 6. Figure \ref{fig:different_K} shows that for $k \geq 3$, performance is comparable, indicating a saturation in performance gains with higher values of $k$. However, the computational cost for training increases polynomially in terms of $k$, because the candidate set $\Candidate_x$ is increasing polynomially. Considering the trade-off between computation and performance, we choose $k=3$ for our experiments.

\section{Related Work}
\label{sec:related}
Recent studies on LLMs has highlighted their ability for ICL, where the model adapts to new tasks solely through inference~\citep{brown2020language}. Recent studies~\citep{rubin2022learning,li2023unified,wang-etal-2024-learning} train dense retrievers based on LLM feedback.
However, those works overlook the interaction between in-context examples. To capture the interplay, \citet{pmlr-v202-ye23c} explain the ICL problem using the determinantal point processes, \citet{zhang-etal-2022-active} also frame ICL problem as a decision making problem but use Q-learning~\citep{DBLP:journals/corr/MnihKSGAWR13}.
Our method presents key advantages over \citet{zhang-etal-2022-active} since we introduce an efficient retrieval head that generates a policy distribution, allowing sample efficiency and to explore diversity, whereas they adopt offline RL and generate an action by evaluating the entire action space (i.e., entire corpus) with the LLM-based Q-network. More differences are listed in the Appendix \ref{sec:appendixB}.

\section{Conclusion}
\label{sec:conclusion}

This paper presents a unified framework that allows LLMs to actively select and rank influential examples, then optimize demonstration selection via reinforcement learning.
This framework enhances the applicability of LLMs to learn what contexts are favorable by the models themselves, paving the way for further similar studies on retrieval-augmented applications.

\section{Limitations}
\label{sec:limitation}
Our proposed sequential demonstration retriever aligns with recent advancements in generative retrieval systems. Unlike generative retrieval, our method does not necessitate supervised fine-tuning for indexing and generating document IDs. However, both methods share a common limitation: the corpus cannot accommodate new entries.

Additionally, computational costs escalate with an increasing number of demonstrations, as we sample a specified quantity at each step. Consequently, our training primarily focuses on a small number of demonstrations.



\bibliography{custom}

\appendix

\section{Experimental Settings}
\label{sec:appendixA}

{\renewcommand{\arraystretch}{1.05}
\begin{table*}[htb]
\small
\centering
\setlength{\abovecaptionskip}{0.2cm}
\setlength{\belowcaptionskip}{-0.4cm}
\setlength\tabcolsep{4.5pt}
\begin{tabular}{lccccccccccc}
\Xhline{1.5pt}
 & SST-2 & MR & CR & ComE & ComV & RTE & SNLI & Reddit & Roc Ending & PHP & DART         \\
\Xhline{1pt} 
Train size                          &1000 &1000 &772 &1000 &1000 &1000 &1000 &1000 &1000 &1000 &1000          \\
Corpus size                       &5000 &5000  &1000 &5000 &5000 &1490 &5000 &5000 &5000 &5000 &5000        \\
Test size                          &872 &2000 &2000 &1000 &1000 &277 &1000 &560 &1000 &1000 &1000         \\
\Xhline{0.8pt}
Evaluation metric &acc &acc &acc &acc &acc &acc &acc &BLEU-1 &BLEU-1 &BLEU-1 &RougeL \\
\bottomrule
\end{tabular}
\caption{Sizes of sampled train set, corpus, and test set that is splitted from each dataset. The term \emph{acc} is accuracy. }
\label{tab:dataset_size} 
\end{table*}
}

\paragraph{Dataset details}
To ensure training efficiency and minimize the impact of varying training and corpus sizes on the RL learning process, we standardize the training examples to 1000 and the corpus size to 5000, except when the original train split is less than 6000. We randomly sample the train set and corpus from each dataset's original training set, ensuring no overlap between them. For inference, limited by computational resources, we randomly sample a test set of 2000 examples. 
Table~\ref{tab:dataset_size} details the sizes of the train set, corpus, and test set, as well as the evaluation metrics.



\section{More Discussion}
\label{sec:appendixB}
Our method offers several key advantages that set it apart from the approach by \citet{zhang-etal-2022-active}:

First, while \citet{zhang-etal-2022-active} they greedily select an action by passing the entire action space (i.e. corpus) to the LLM-based Q-network, our method introduces an efficient retrieval head to generate a policy distribution. This allows for sampling that enhances diversity, unlike their off-line RL approach which fails to explore diverse options.

Second, our initial retrieval head demonstrates retrieval performance on par with dense retrievers, achieving sample efficiency and optimizing within a small subset of representative examples. In contrast, their initial Q-network approximates from random selection.

Third, \citet{zhang-etal-2022-active} do not include a reward model to produce stable reward signal, which we have proven necessary in our experiment section.

\section{Additional Experiments}
\label{sec:appendixC}
\subsection{Transferability of the learned policy}
{\renewcommand{\arraystretch}{1.05}
\begin{table}[h]
\small
\centering
\setlength{\abovecaptionskip}{0.2cm}
\setlength{\belowcaptionskip}{-0.4cm}
\setlength\tabcolsep{4.5pt}
\begin{tabular}{lc}
\Xhline{1.5pt}
 & RTE    \\
\Xhline{1pt} 
Llama-3 RL retriever $\rightarrow$ Llama-3 ICL (self-ICL)     &75.81  \\ 
Llama-2 initial retriever $\rightarrow$ Llama-2 ICL           &66.78   \\
Llama-3 RL retriever $\rightarrow$ Llama-2 ICL                &74.36   \\
\bottomrule  
\end{tabular}
\caption{Transferability across LLM.}
\label{tab:transfer}
\end{table}
}
We investigate the transferability of a policy learned through our proposed method to a different large language model (LLM). Initially, we optimize the retrieval head in conjunction with a frozen Llama-3 model. After the reinforcement learning process, the trained retrieval head generates an optimal policy that selects the most appropriate demonstrations for a given input. Subsequently, we provide these demonstrations to a different LLM, such as Llama-2, to assess the effectiveness of the transferred demonstrations.
This transfer setting, referred to as \textbf{Llama-3 RL retriever $\rightarrow$ Llama-2 ICL}, is presented in the last row of Table~\ref{tab:transfer}.
The baseline \textbf{Llama-2 initial retriever $\rightarrow$ Llama-2 ICL } employs Llama-2 for both retrieval head initiation and ICL execution. Results demonstrate that the transferred retriever enhances performance compared to the baseline, confirming the efficacy of our self-ICL approach in training transferable retrievers.

\section{Example Cases of Tasks}
\label{sec:appendixD}

\clearpage
\onecolumn
\begin{longtable}{p{0.95\textwidth}}
\label{tab:prompts_clf}\\

\toprule[2pt]
\endfirsthead

\toprule
\endhead

\multicolumn{1}{c}{\textbf{Datasets}: SST2} \\ \midrule[0.2pt]
\textbf{Task}: Sentiment Analysis \\
\textbf{Label Verbalizer}: {Positive, Negative}\\
\midrule[0.2pt]
\multicolumn{1}{c}{Example}\\
\midrule[0.2pt]
Text:there 's no indication that he 's been responsible for putting together any movies of particular value or merit\\
Sentiment "Positive" or "Negative"?Negative\\
Text:, but i believe a movie can be mindless without being the peak of all things insipid \\
Sentiment "Positive" or "Negative"?Negative\\
Text:deftly setting off uproarious humor with an underlying seriousness that sneaks up on the viewer , providing an experience that is richer than anticipated . \\
Sentiment "Positive" or "Negative"?Positive\\
Text:the movie has an infectious exuberance that will engage anyone with a passing interest in the skate/surf culture , the l.a. beach scene and the imaginative ( and sometimes illegal ) ways kids can make a playground out of the refuse of adults . \\
Sentiment "Positive" or "Negative"?\\

\\ \midrule[2pt]
\multicolumn{1}{c}{\textbf{Datasets}: MR} \\ \midrule[0.2pt]
\textbf{Task}: Sentiment Analysis \\
\textbf{Label Verbalizer}: {Positive, Negative}\\
\midrule[0.2pt]
\multicolumn{1}{c}{Example}\\
\midrule[0.2pt]
Text:murderous maids pulls no punches in its depiction of the lives of the papin sister and the events that led to their notorious rise to infamy . . .\\
Sentiment "Positive" or "Negative"?Positive\\
Text:witty , contemplative , and sublimely beautiful .\\
Sentiment "Positive" or "Negative"?Positive\\
Text:a compelling film .\\
Sentiment "Positive" or "Negative"?Positive\\
Text: . . . a joke at once flaky and resonant , lightweight and bizarrely original .\\
Sentiment "Positive" or "Negative"?

\\ \midrule[2pt]
\multicolumn{1}{c}{\textbf{Datasets}: CR} \\ \midrule[0.2pt]
\textbf{Task}: Sentiment Analysis \\
\textbf{Label Verbalizer}: {Positive, Negative}\\
\midrule[0.2pt]
\multicolumn{1}{c}{Example}\\
\midrule[0.2pt]
Text:the navigation takes so much time that it would eventually drive you crazy .\\
Sentiment "Positive" or "Negative"?Negative\\
Text:even with that , i highly recommend this router - outstanding performer .\\
Sentiment "Positive" or "Negative"?Positive\\
Text:installation was as near automatic as can be .\\
Sentiment "Positive" or "Negative"?Positive\\
Text:i did not want to have high expectations for this apex player because of the price but it is definitely working out much better than what i would expect from an expensive high-end player .\\
Sentiment "Positive" or "Negative"?

\\ \midrule[2pt]
\multicolumn{1}{c}{\textbf{Datasets}: ComE} \\ \midrule[0.2pt]
\textbf{Task}: Validation and Explanation \\
\textbf{Label Verbalizer}: {A, B, C}\\
\midrule[0.2pt]
\multicolumn{1}{c}{Example}\\
\midrule[0.2pt]
Select the most corresponding reason why this statement is against common sense. The lighthouse misled the ship. Options: A. The status of the lighthouse is very important to the ship. B. At night, the lighthouse will show the way home for the ship. C. Ships can sail on the sea under the guides of the lighthouse\\
Answer:B. At night, the lighthouse will show the way home for the ship.\\
Select the most corresponding reason why this statement is against common sense. We can always see stars when we look up at night Options: A. We cannot see stars if the weather is not good B. Looking at stars with your lover is a romantic thing C. We may see the moon when we look up at night\\
Answer:A. We cannot see stars if the weather is not good\\
Select the most corresponding reason why this statement is against common sense. My dog loves reading books. Options: A. Dogs love eating books. B. Dogs can't reading. C. Dogs don't play with books.\\
Answer:B. Dogs can't reading.\\
Select the most corresponding reason why this statement is against common sense. Roberts' room is sleeping Options: A. A room cannot close his eyes, because he has no eyes, so he can't sleep. B. Robert won't let the room sleep because he needs to sleep in it first. C. Robert can sleep in his room\\
Answer:

\\ \midrule[2pt]
\multicolumn{1}{c}{\textbf{Datasets}: ComV} \\ \midrule[0.2pt]
\textbf{Task}: Validation and Explanation\\
\textbf{Label Verbalizer}: {Statement 1, Statement 2}\\
\midrule[0.2pt]
\multicolumn{1}{c}{Example}\\
\midrule[0.2pt]
Which statement of the two is against common sense? Statement 1: he has ten toes on one foot Statement 2: he has ten toes on his feet\\
Answer:Statement 1: he has ten toes on one foot\\
Which statement of the two is against common sense? Statement 1: Kate slept on the bed last night Statement 2: Kate slept on the ceiling last night\\
Answer:Statement 2: Kate slept on the ceiling last night\\
Which statement of the two is against common sense? Statement 1: he sits on the sofa Statement 2: he sits on a lake\\
Answer:Statement 2: he sits on a lake\\
Which statement of the two is against common sense? Statement 1: Roberts' room is sleeping Statement 2: Robert close he's eyes and he is sleeping in his room\\
Answer:

\\ \midrule[2pt]
\multicolumn{1}{c}{\textbf{Datasets}: RTE} \\\midrule[0.2pt]
\textbf{Task}: Natural Language Inference \\
\textbf{Label Verbalizer}: {True, False}\\
\midrule[0.2pt]
\multicolumn{1}{c}{Example}\\
\midrule[0.2pt]
Text:Tony Blair, former Prime Minister of the United Kingdom, has left the Church of England and joined the Roman Catholic Church. Blair, currently the special envoy for Quartet on the Middle East, has long been attending mass with his wife and four children, who are all Catholic. Cardinal Cormac Murphy-O'Connor received Blair into full communion with the Catholic Church during Mass at Archbishop's House, Westminster, on Friday. Question: Blair belongs to the Church of England. True or False?\\
Answer:False\\
Text:Newspapers choke on rising paper costs and falling revenue. Question: The cost of paper is rising. True or False?\\
Answer:True\\
Text:His family has steadfastly denied the charges. Question: The charges were denied by his family. True or False?\\
Answer:True\\
Text:Deceased U.S. soldiers and their effects were evacuated to Japan and then shipped home in refrigerated containers for interment in the U.S. Question: The U.S. military evacuated U.S. citizens. True or False?\\
Answer:

\\ \midrule[2pt]
\multicolumn{1}{c}{\textbf{Datasets}: SNLI} \\\midrule[0.2pt]
\textbf{Task}: Natural Language Inference \\
\textbf{Label Verbalizer}: {Entailment, Contradiction, Inconclusive}\\
\midrule[0.2pt]
\multicolumn{1}{c}{Example}\\
\midrule[0.2pt]
Three males are drinking beer. Based on that information, is the claim A small group of guys is drinking "Entailment", "Contradiction", or "Inconclusive"?\\
Answer:Entailment\\
A man is laughing at a bar drinking a beer. Based on that information, is the claim A person is chuckling at a bar drinking a alcoholic beverage "Entailment", "Contradiction", or "Inconclusive"?\\
Answer:Entailment\\
A brown dog is soaked and is walking out of the water. Based on that information, is the claim A cat is soaked and is walking out of the water "Entailment", "Contradiction", or "Inconclusive"?\\
Answer:Contradiction\\
People are throwing tomatoes at each other. Based on that information, is the claim The people are sitting and eating their food "Entailment", "Contradiction", or "Inconclusive"?\\
Answer:

\\ \midrule[2pt]
\multicolumn{1}{c}{\textbf{Datasets}: Reddit} \\\midrule[0.2pt]
\textbf{Task}: Text Summarization \\
\midrule[0.2pt]
\multicolumn{1}{c}{Example}\\
\midrule[0.2pt]
Summarize the text:so in my group we have this joke about the yoghurt called chobani because whenever some brings it it always goes over all there clothes . so one day i was opening my chobani and everyone started running away because they though that i was going to smash it so it went on someone . so i started to be a dick and pretend to smash it but i accidentally hit it and the top of the yoghurt flicked up and all of the yoghurt went all over my face and everyone started laughing at me . it also went all over my uniform so i got so many weird looks because i had a massive white stain on my shirt\\
TL;DR:i was being a dick pretending to put yoghurt on my friends and it back fired and went all over me\\
Summarize the text:i decided to go out although i was extremely tired . i head dt for a dj set at the local club . ive drank a bit and taken a bit of m ... . so im in good spirits . i head down to meet a friend . when i get in there some random guy ( someone i recognize though ) gives me some fudge . now this is where you should be able to go back to basics as a kid ... dont take candy/food from a stranger . also ... its in a club , what the fuck did i think ? ? ? the fudge was wrapped and all , so i didnt think anything crazy . well i was in a happy zone and didnt think anything of it , and devoured it . man it was good to . i later saw a lot of joints out , and i didnt connect the dots until that moment ... . that wasnt just some fudge ... . that fudge was packed ... . and i knew once i digested it i was going to be a different state of mind . about 1.5hrs later , it hits ... . im fucked ... ... i feel like i put space goggles on because visually i was gone pretty much ... i had a decent time overall , but would of preferred just sticking with my other 2 uppers and not the fudge .\\
TL;DR:took fudge from stranger , didnt think about it , ate it ... . not just fudge .\\
Summarize the text:long time lurker first time poster hapened awhile ago typing this on my phone all that jazz so i am a 15 yo guy and i really want a job you know because i like money and my mom is having a hard time ... anyway i see on a website that this mt was hiring at 15. checked the website , and they have a slot in the one near me so i happily apply online and i get a call on my phone , `` is this so and so i was calling about your application and wanted to check that all your credentials were right '' me being a stupid 15 year old disney realise the call was n't from my area , or state and i confirmed my credentials '' ok we will send you applications on email thanks '' sweet i think then the fuck up i start getting calls non stop i have been called 9-10 times a day by indian people wanting to get me jobs every time a block 1 a new 1 pops up it got to the point a teacher took my phone in class because i got 2 calls in 45 min . yeah so i do n't want to change my number because i just moved and might not be able to talk to my old friends , and i 'm having a bad time\\
TL;DR:applied for a movie theatre got calls from everyone else\\
Summarize the text:as per often on this subreddit , this did n't happen today , but last night at around 11:30. this is my first post ever , but definitely not my first fuck up . i was bored and watching the usual chemistry videos as i do almost every day . i came across one where you could make chloroform with bleach and acetone . it seemed easy because simply mixing them would cause the reaction to occur . after watching the video and seeing that it should be pretty quick for it to react ; i decided to try it myself . i quickly scurried to the basement and started searching for the acetone , after 15 minutes of searching i decided to give up because it was nowhere to be found , until i sat down and noticed the orange container of acetone right next to me . being so excited about finding it i quickly poured a little bit into a *beaker* ( a small glass bowl ) and rushed to get the bleach . after i added a little bit of the bleach i mixed it and waited for the heat from the exothermic reaction to take place . after about 3 minutes of waiting for heat and an immiscible substance to form i gave up and added more acetone hoping it would eventually work . after another minute.. still nothing . another minute later i was adding more bleach hoping that it would work . the fu was when i was so determined to have a chemical reaction happen . i gave up trying to make chloroform so i decided to add some hydrochloric acid to the mixture . bored and all and knowing it would make chlorine gas , i decided to do it anyways . after i mixed it , nothing happened again , so my smart ass decided to smell it instead of whiffing it like you 're supposed to . as soon as i did that i felt my nose burn and i started coughing . i ran into the bathroom turned on the fan that sucks air and takes it outside and flushed the mixture down the toilet . all in all it was n't too bad except for me being dizzy the rest of the night and almost falling down the stairs .\\
TL;DR:

\\ \midrule[2pt]
\multicolumn{1}{c}{\textbf{Datasets}: Roc Ending} \\\midrule[0.2pt]
\textbf{Task}: Story Generation \\
\midrule[0.2pt]
\multicolumn{1}{c}{Example}\\
\midrule[0.2pt]
Unfinished story:I loved potato chips. I wanted to create my own flavor of chips. I decided to make some beer flavored chips. Alas, they were disgusting.\\
End of the story:I realized I should let the professionals stick to making chips.\\
Unfinished story:Hallee and her friends are excited there is a big snowstorm coming. While they are in school, it starts to snow. Their teacher tells them there will be no school the next day. Hallee and her friends are so excited for a snow day the next day!\\
End of the story:After school, they go outside and play in the snow!\\
Unfinished story:Jake wanted to marry his girlfriend of 5 Years. He saved up for a wedding ring. Jake proposed to her at dinner. Her girlfriend said yes.\\
End of the story:Jake kissed his girlfriend afterwards.\\
Unfinished story:Ruby was poor but always tried to buy a weekly luxury. One week she decided to buy a lottery ticket. She watched the drawings on TV that night, excited. The large jackpot was a million, and she didn't get it.\\
End of the story:

\end{longtable}
\twocolumn

\end{document}